\documentclass[10pt,twocolumn,letterpaper]{article}

\usepackage[pagenumbers]{cvpr} 
\usepackage{pifont}
\usepackage[dvipsnames]{xcolor}

\newcommand{\zzx}[1]{{ #1}}
\newcommand{\shortname}{{\emph{TGS}}}

\definecolor{cvprblue}{rgb}{0.21,0.49,0.74}
\usepackage[pagebackref,breaklinks,colorlinks,citecolor=cvprblue]{hyperref}

\def\confName{CVPR}
\def\confYear{2024}

\title{Triplane Meets Gaussian Splatting:\\Fast and Generalizable Single-View 3D Reconstruction with Transformers}

\author{Zi-Xin Zou$^{1}$ \quad Zhipeng Yu$^{2}$ \quad Yuan-Chen Guo$^{1,2*}$ \quad Yangguang Li$^{2}$ \quad \\ Ding Liang$^{2}$ \quad Yan-Pei Cao$^{2\dag}$ \quad
Song-Hai Zhang$^{1\dag}$ \\
~\\
$^1$BNRist, Tsinghua University \quad $^2$VAST \\
~\\
}

\begin{document}
\twocolumn[{%
\renewcommand\twocolumn[1][]{#1}%
\maketitle
\begin{center}
    \centering
    \captionsetup{type=figure}
    \includegraphics[width=\linewidth]{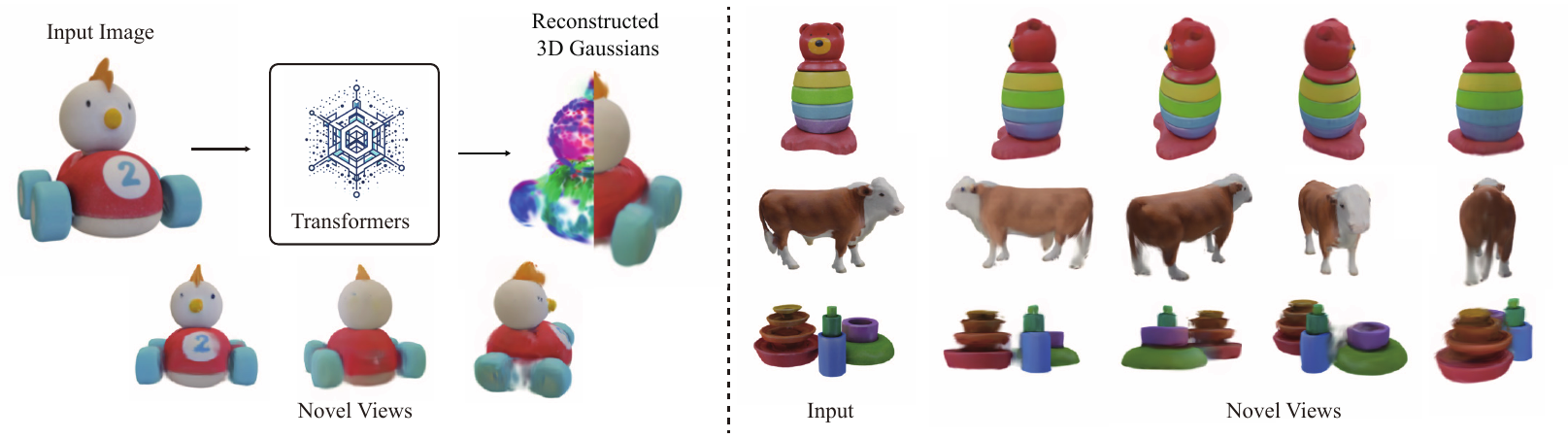}
    \captionof{figure}{We propose a method that enables fast reconstruction from a single-view image. We build the 3D representation upon a hybrid Triplane-Gaussian representation by evaluating a transformer-based framework, from which 3D Gaussians would be decoded (Left). Based on this, we can perform novel view synthesis with high quality and fast rendering speed via Gaussian Splatting (Right).}
    \label{fig:teaser}
\end{center}%
}]
{\let\thefootnote\relax\footnote{{$^*$ Intern at VAST}}}
{\let\thefootnote\relax\footnote{{$^\dag$ Corresponding authors}}}
\begin{abstract}
Recent advancements in 3D reconstruction from single images have been driven by the evolution of generative models. Prominent among these are methods based on Score Distillation Sampling (SDS) and the adaptation of diffusion models in the 3D domain.
Despite their progress, these techniques often face limitations due to slow optimization or rendering processes, leading to extensive training and optimization times.
In this paper, we introduce a novel approach for single-view reconstruction that efficiently generates a 3D model from a single image via feed-forward inference. Our method utilizes two transformer-based networks, namely a point decoder and a triplane decoder, to reconstruct 3D objects using a hybrid Triplane-Gaussian intermediate representation.
This hybrid representation strikes a balance, achieving a faster rendering speed compared to implicit representations while simultaneously delivering superior rendering quality than explicit representations.
The point decoder is designed for generating point clouds from single images, offering an explicit representation which is then utilized by the triplane decoder to query Gaussian features for each point. This design choice addresses the challenges associated with directly regressing explicit 3D Gaussian attributes characterized by their non-structural nature.
Subsequently, the 3D Gaussians are decoded by an MLP to enable rapid rendering through splatting.
Both decoders are built upon a scalable, transformer-based architecture and have been efficiently trained on large-scale 3D datasets. 
The evaluations conducted on both synthetic datasets and real-world images demonstrate that our method not only achieves higher quality but also ensures a faster runtime in comparison to previous state-of-the-art techniques.
Please see our project page at \href{https://zouzx.github.io/TriplaneGaussian/}{https://zouzx.github.io/TriplaneGaussian/}
\end{abstract}

\section{Introduction}
Digitizing 3D objects from single 2D images represents a crucial and longstanding challenge in both computer vision and graphics. The significance of this problem stems from its broad applications, notably in augmented reality (AR) and virtual reality (VR). However, the inherent ambiguity and lack of information in single images pose a substantial challenge in accurately recovering the complete, high-quality shape and texture of an object from such a constrained perspective.

The recent surge in image generation using diffusion models~\cite{DBLP:conf/icml/RameshPGGVRCS21,DBLP:conf/cvpr/RombachBLEO22} enabled approaches to ``imagine'' images from novel viewpoints by leveraging these models with pose transformation conditions~\cite{DBLP:journals/corr/abs-2303-11328} or novel view rendering feature maps~\cite{DBLP:journals/corr/abs-2304-02602,DBLP:conf/cvpr/ZhouT23}.
Despite these advancements, achieving consistent novel view synthesis remains a complex task, largely due to the lack of 3D structural constraints. 
 Recent efforts incorporating multi-view attention or 3D-aware feature attention for simultaneous multi-view generation~\cite{DBLP:journals/corr/abs-2308-16512,DBLP:journals/corr/abs-2309-03453} have not fully mitigated this challenge.
Furthermore, these methods require a time-intensive, object-specific optimization process to distill diffusion priors~\cite{DBLP:conf/iclr/PooleJBM23,DBLP:conf/cvpr/Lin0TTZHKF0L23,DBLP:conf/cvpr/Melas-KyriaziL023,Tang_2023_ICCV,Chen_2023_ICCV}, combined with monocular cues (depth or normals~\cite{DBLP:conf/iccv/EftekharSMZ21,DBLP:conf/iccv/LongLL0TYW21}), into neural 3D representations such as NeRF~\cite{DBLP:conf/eccv/MildenhallSTBRN20} and DMTet~\cite{DBLP:conf/nips/ShenGYLF21}.
This optimization procedure is often too slow for practical applications requiring rapid 3D creation.

Instead of optimizing a 3D representation, alternative approaches utilize category-specific shape priors to regress 3D representations (e.g., point clouds~\cite{DBLP:conf/cvpr/FanSG17,DBLP:conf/cvpr/WuZXZC20}, voxels~\cite{DBLP:conf/eccv/GirdharFRG16,DBLP:conf/nips/0001WXSFT17}, meshes~\cite{DBLP:conf/cvpr/WorchelDHSFE22,DBLP:conf/eccv/WangZLFLJ18}, and NeRF~\cite{DBLP:conf/cvpr/YuYTK21}) from single images.
Among various 3D representations, the triplane representation is notable for its compactness and efficient expressiveness, with its implicit nature facilitating the learning of 3D structures via volume rendering~\cite{DBLP:conf/eccv/MildenhallSTBRN20}.
Nevertheless, the inherent complexity of volume rendering imposes significant runtime and memory costs, adversely impacting training efficiency and real-time rendering capabilities.
Recently, Gaussian Splatting~\cite{DBLP:journals/tog/KerblKLD23} has made strides by using anisotropic 3D Gaussians for high-quality radiance field representation, enhancing both novel view synthesis quality and rendering speed through adaptive density control and efficient, differentiable rendering.
Yet, the direct learning of 3D Gaussian representations from images presents a substantial challenge due to their discrete, non-structural, and higher-dimensional nature compared to implicit representations that only require decoding RGB and density. Additionally, the non-structural and explicit characteristics of 3D Gaussians further complicate the learning process.  (see the first image of Figure~\ref{fig:ab}).
Consequently, developing an effective method to unleash the potential of Gaussian Splatting for generalizable 3D learning remains a significant area for exploration.

In this work, we introduce \shortname, a novel 3D reconstruction approach from single-view images, leveraging the strengths of \textbf{T}riplane and \textbf{G}aussian \textbf{S}platting.
Our method employs a hybrid explicit-and-implicit 3D representation, facilitating fast and high-quality 3D reconstruction and novel view synthesis.
We diverge from traditional 3D representation methods by proposing a hybrid Triplane-Gaussian model, which marries an explicit point cloud with an implicit triplane field.
The explicit point cloud outlines the object's rough geometry, while the implicit triplane field refines the geometry while encoding 3D Gaussian properties such as opacity and spherical harmonics.
This separation of position (explicit) and Gaussian properties (implicit) significantly enhances the quality of 3D reconstruction, while maintaining the efficiency of Gaussian Splatting.
Based on such representation, our approach includes two networks for reconstructing the point cloud and triplane from the input image, respectively.
We employ a fully transformer-based architecture for both the point cloud and triplane networks, enabling the interaction of latent features with input image features through cross-attention.
This transformer design ensures scalability and supports large-scale, category-agnostic training, thereby enhancing the model's generalizability to real-world objects. 
Moreover, we augment each 3D Gaussian with local image features by projecting explicit geometry onto the input image plane. 
Overall, the proposed approach allows for fast, high-resolution rendering and facilitates efficient end-to-end training of the entire pipeline.

To the best of our knowledge, this is the first study to achieve generalizable 3D reconstruction from single-view images using Gaussian Splatting, opening new possibilities in fast 3D content creation and rendering.
Our extensive experiments on the Google Scanned Object (GSO) dataset~\cite{DBLP:conf/icra/DownsFKKHRMV22} demonstrate that our method surpasses existing baselines in both geometry reconstruction and novel-view synthesis. Owing to its feed-forward model and the hybrid representation, our approach achieves object reconstruction in a matter of seconds, significantly faster than prior methods.
We summarize our main contributions as follows:
\begin{itemize}
    \item We propose Triplane-Gaussian, a new hybrid representation that leverages both explicit and implicit representation for fast and high-quality single-view reconstruction.
    \item By leveraging local image features for projection-aware conditioning in the transformer networks, we achieve greater consistency with the input observation.
    \item Experiments show that our approach outperforms state-of-the-art baselines in terms of reconstruction quality and speed.
\end{itemize}

\section{Related Works}
\begin{figure*}[t]
    \centering
    \includegraphics[width=1.0\linewidth]{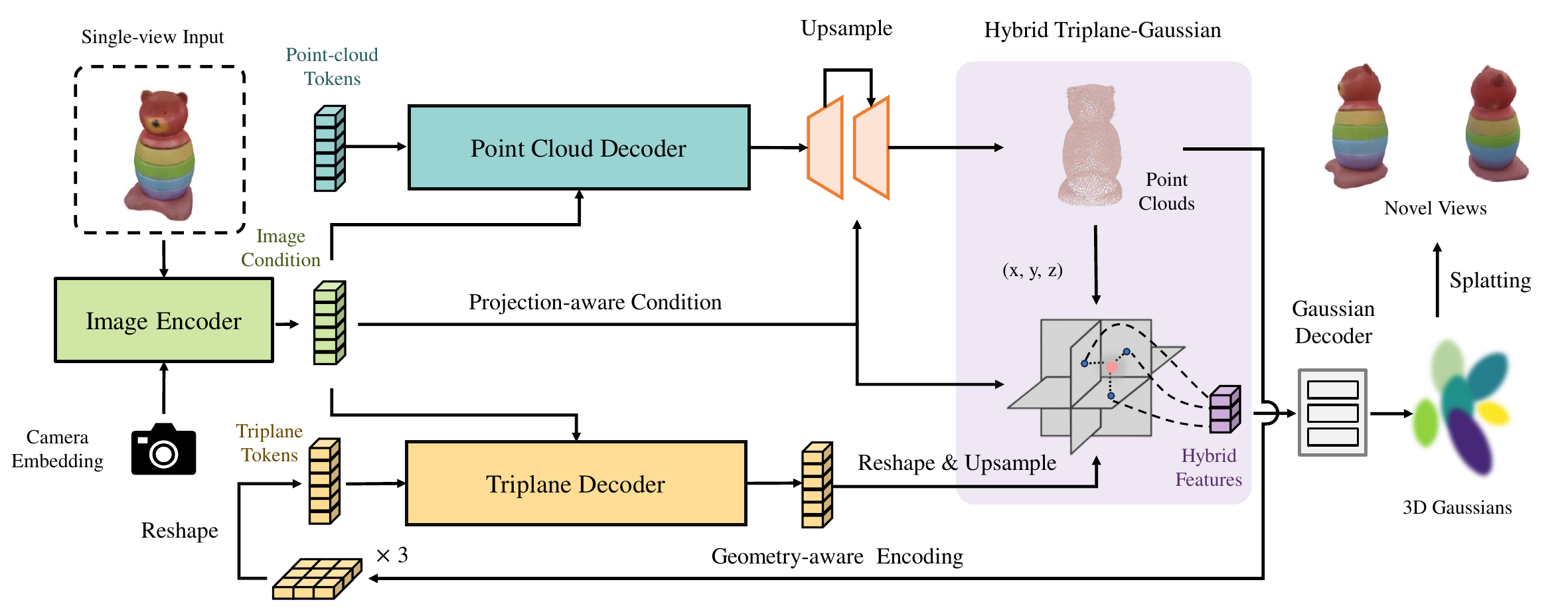}
    \caption{\textbf{The overview of our framework}. Given an image with its camera parameters, we first encode them into a set of latent feature tokens by leveraging a pre-trained ViT model. Our two transformer-based networks, point cloud decoder and triplane decoder, take initial positional embedding as input and project image tokens onto latent feature tokens of respective 3D representation via cross-attention. 
    Subsequently, a point cloud and a triplane can be de-tokenized from the output of decoders, respectively. 
    After the point cloud decoder, we adapt a point upsampling module~\cite{xiang2023SPD,xiang2021snowflakenet} with condition-aware projection to densify the point cloud.
    Additionally, we utilize a geometry-aware encoding to project point cloud features into the initial positional embedding of triplane latent.
    Finally, 3D Gaussians are decoded by the point cloud, the triplane features and image features for novel view rendering.}
    \label{fig:overview}
    \vspace{-15pt}
\end{figure*}
\vspace{-2pt}
\paragraph{Single View 3D Reconstruction.}
3D reconstruction from single view image is an ill-posed problem that garnered substantial attention from many researchers, due to the lack of geometry cues from single view.
Most studies learn strong 3D priors from 3D synthetic models~\cite{DBLP:journals/corr/ChangFGHHLSSSSX15} or real scans~\cite{DBLP:conf/cvpr/DaiCSHFN17}.
Previous works have constructed priors based on collections of 3D primitives using an encoder-decoder approach~\cite{DBLP:conf/nips/XuWCMN19,DBLP:conf/cvpr/MeschederONNG19}.
For example, they utilize image encoder to encode visual information into compact latent features (e.g., vectors and volumes), and the 3D model can be decoded from the features for specific representation.
However, these methods often are limited to specific category.
And most of them focus on geometry reconstruction and have poor performance on texture.
Recently, transformer architecture~\cite{DBLP:conf/nips/VaswaniSPUJGKP17} has shown its capacity and scalability in the field of computer vision~\cite{DBLP:conf/iclr/DosovitskiyB0WZ21,DBLP:conf/iccv/CaronTMJMBJ21,DBLP:conf/icml/RadfordKHRGASAM21,DBLP:journals/corr/abs-2204-06125} and other areas~\cite{DBLP:conf/naacl/DevlinCLT19,DBLP:conf/nips/BrownMRSKDNSSAA20}.
GINA-3D~\cite{DBLP:conf/cvpr/ShenYQNDGZA23} and LRM~\cite{DBLP:journals/corr/abs-2311-04400} propose a full transformer-based pipeline to decode a NeRF representation from triplane features.
MCC~\cite{DBLP:conf/cvpr/Wu0MFG23} also trains a transformer decoder to predict occupancy and color from RGB-D inputs.
In this work, we adopt the scalable transformer-based architecture to learn generalizable 3D representation from single image.

\vspace{-10pt}
\paragraph{3D Generation conditioned on image.}
Contrary to learning 3D geometry prior and reconstructing an object from image in a feed-forward manner, some recent works treat this task as a conditional generation. They aim to exploit techniques such as GAN~\cite{DBLP:conf/nips/0004SWCYLLGF22}, diffusion model~\cite{DBLP:journals/tog/ZhengPWTLS23,DBLP:conf/cvpr/Melas-Kyriazi0V23,DBLP:conf/cvpr/KarnewarVNM23,DBLP:conf/cvpr/ChengLTSG23,DBLP:conf/nips/zengVWGLFK22} to sample a high 3D content from pre-trained distribution.
Point-E~\cite{DBLP:journals/corr/abs-2212-08751} and Shap-E~\cite{DBLP:journals/corr/abs-2305-02463} are trained on several million 3D models to generate point clouds and parameters of implicit function (e.g. neural radiance field).
However, these 3D native generative models are limited due to lack of a substantial mount of high-quality 3D model datasets.

Inspired from success of 2D generative models (e.g. DALL-E~\cite{DBLP:conf/icml/RameshPGGVRCS21} and Stable Diffusion~\cite{DBLP:conf/cvpr/RombachBLEO22}) with impressive and high-quality image generation, a large number of studies~\cite{DBLP:conf/iclr/PooleJBM23,DBLP:conf/cvpr/Lin0TTZHKF0L23,Chen_2023_ICCV,DBLP:conf/cvpr/Melas-KyriaziL023} incorporate 2D diffusion and CLIP priors into per-shape optimization via various score distillation sampling strategy~\cite{DBLP:conf/iclr/PooleJBM23,DBLP:conf/cvpr/WangDLYS23,DBLP:journals/corr/abs-2305-16213}, or with other monocular prior losses.
Most off-the-shelf high-quality text-to-image diffusion models cannot perfect generate images at given novel view that plausibly constitute other views of the given object.
Alternatively, some works explore view-conditioned diffusion to learn control mechanisms that manipulate the camera viewpoint in large-scale diffusion models~\cite{DBLP:journals/corr/abs-2303-11328}.
However, these methods necessitate a time-consuming per-shape optimization for different 3D representations (i.e., NeRF, mesh, 3D Gaussian, SMPL human model) via differentiable rendering.
Some other works improves the multi-view consistency across different novel views~\cite{DBLP:journals/corr/abs-2308-16512,DBLP:journals/corr/abs-2309-03453}, which directly enables NeRF optimization with pixel-wise loss with distillation, achieving impressive performance in few minutes.
One-2-3-45~\cite{liu2023one} proposes to combine the 2D generative models and mutli-view 3D reconstruction, leading to superiority on both quality and efficiency.

\vspace{-10pt}
\paragraph{Neural 3D Representation and Rendering.}
With the recent advancement of neural networks, neural implicit representations~\cite{DBLP:conf/cvpr/ParkFSNL19,DBLP:conf/cvpr/MeschederONNG19,DBLP:conf/cvpr/JiangSMHNF20}, especially neural radiance field (NeRF)~\cite{DBLP:conf/eccv/MildenhallSTBRN20} with its variants~\cite{DBLP:conf/iccv/BarronMTHMS21,DBLP:conf/nips/LiuGLCT20,DBLP:conf/cvpr/0007CS23,DBLP:conf/cvpr/ZhangBS0X22}, have shown its promising performance in novel-view synthesis and 3D reconstruction, only using dense images via differentiable volume rendering.
Despite many efforts dedicated to enhancing the rendering speed of neural implicit representations~\cite{DBLP:journals/tog/MullerESK22,DBLP:conf/iccv/GarbinK0SV21}, their efficiency still falls short when compared with point-based rendering~\cite{DBLP:conf/cvpr/LassnerZ21,DBLP:journals/cgf/KopanasPLD21}.
Certain works enhance points with neural features and employ a UNet~\cite{DBLP:conf/eccv/AlievSKUL20,DBLP:conf/cvpr/RakhimovALB22} for fast or even real-time rendering, leveraging its advantages of fast GPU/CUDA-based rasterization.
These methods, relying on 2D CNN thus lacking 3D awareness, often occur over- or under-reconstruction in hard cases (e.g. featureless areas or thin structures).
The 3DGS~\cite{DBLP:journals/tog/KerblKLD23}, as a special point-based rendering, stands out for its distinctive utilization of explicit representation and differential point-based splatting techniques, thereby facilitating real-time rendering at novel views.
However, it's difficult to directly regress 3D Gaussians from image for high-quality novel view synthesis.
In this paper, we leverage a hybrid 3D representation, facilitating fast and generalizable reconstruction and rendering.

\section{Method}
In the subsequent sections, we present our approach for 3D object reconstruction from single-view images.
We introduce a new hybrid 3D representation that combines explicit point cloud geometry and implicit triplane features,
allowing for efficient rendering without compromising on quality (Section~\ref{subsec:repr}).
In order to optimize the hybrid representation from a singe-view input, we first employ a transformer-based point cloud decoder to predict coarse points from image features
and upsample the coarse points to a dense point cloud.
Subsequently, a triplane decoder takes these points along with the image features and outputs the triplane features. 
Finally, the hybrid Triplane-Gaussian will encompass point clouds, triplane features, and image features to generate representative hybrid features for 3D Gaussian attributes decoding (Section~\ref{subsec:rec}).
Benefiting from differentiable splatting, we can train the full framework end-to-end on large-scale category-agnostic datasets efficiently (Section~\ref{subsec:train}).

\subsection{Hybrid Triplane-Gaussian}\label{subsec:repr}
Gaussian Splatting, while offering advantages such as high-quality rendering and speed, remains unexplored in its application to generalizable reconstruction settings.
A preliminary approach might consider treating 3D Gaussians as an explicit point cloud enriched with specific attributes. One possibility involves adapting existing point cloud reconstruction or generation models~\cite{DBLP:journals/corr/abs-2212-08751,DBLP:conf/cvpr/Melas-Kyriazi0V23} to include additional 3D Gaussian attributes besides point positions.
Pursuing this concept, we initially developed a 3D Gaussian reconstruction model based on a transformer architecture aimed at directly predicting 3D Gaussians. 
However, as shown in the experiment in Section \ref{subsec:ab}, it fails to generate satisfied 3D objects from a single image.
This shortfall might be attributed to the discrete, non-structural, and high-dimensional nature of 3D Gaussians, which complicates the learning process.
In response, we introduce \emph{Triplane-Gaussian}, a new hybrid 3D representation that merges the benefits of both triplane and point cloud approaches for 3D Gaussian representation.
This hybrid model offers fast rendering, high quality, and generalizability, as depicted in Figure~\ref{fig:ab}.

The hybrid representation includes a point cloud $P \in R^{N \times 3}$ which provides explicit geometry, and a triplane $T \in R^{3 \times C \times H \times W}$ which encodes an implicit feature field, where 3D Gaussian attributes can be decoded.
The triplane $T$ comprises three axis-aligned orthogonal feature planes $\{T_{\mathrm{xy}},T_{\mathrm{xz}},T_{\mathrm{yz}}\}$.
For any position $\mathbf{x}$, we can query the corresponding feature vector from triplane by projecting it onto axis-aligned feature planes, and concatenate three trilinear interpolated features as the final feature $f_{t}=interp(T_{\mathrm{xy}}, p_{\mathrm{xy}})\oplus interp(T_{\mathrm{xz}}, p_{\mathrm{xz}})\oplus interp(T_{\mathrm{yz}}, p_{\mathrm{yz}})$,  where $interp$ and $\oplus$ denote the trilinear interpolation and concatenation operation, and $p$ denotes the projected position on each plane.

\vspace{-10pt}
\paragraph{3D Gaussian Decoder.}
For a given position $\mathbf{x} \in R^3$ from point cloud $P$, we query feature $f$ from the triplane and adopt an MLP $\phi_{g}$ to decode the attributes of the 3D Gaussians derived from the point cloud. 
The 3D Gaussians' attributes include opacity $\alpha$, anisotropic covariance (represented by a scale $s$ and rotation $q$) and spherical harmonics (SH) coefficients $sh$ ~\cite{DBLP:journals/tog/KerblKLD23}:
\begin{align}
    (\Delta \mathbf{x}', \alpha, s, q, sh) = \phi_{g}(\mathbf{x}, f) \label{equ:gs_decode}
\end{align}
Since surface points are not always the best choice for the 3D Gaussian representation, we additionally predict a small offset $\Delta \mathbf{x}$ for the position, and the final position is computed by $\mathbf{x}=\mathbf{x}+\Delta \mathbf{x}$.

Regarding the single-view 3D reconstruction task we focus on in this paper, we additionally augmented the queried triplane features with projected image features as in PC$^{2}$~\cite{DBLP:conf/cvpr/Melas-Kyriazi0V23} to fully utilize the local features from the input image. 
Specifically, we concatenate the triplane feature $f_t$ with projected local features $f_l$ from explicit geometry as $f$ in Equation~\ref{equ:gs_decode}. 
Given an input camera pose $\pi$ and a point cloud $P$, the local projection feature can be calculated by the projection function $\mathcal{P}$, where $f_l=\mathcal{P}(\pi, P)$.
Following~\cite{DBLP:conf/cvpr/Melas-Kyriazi0V23}, we consider the self-occlusion problem of the point cloud and implement the projection by point rasterization.
The local features include the RGB color, DINOv2~\cite{dinov2} feature, mask, and 2-dimensional distance transform corresponding to the mask region.
In this way, the projection-aware condition can improve the texture quality under the input viewing angle.

\vspace{-10pt}
\paragraph{Rendering.}
Based on such representation, we can perform image rendering by utilizing the efficient Gaussian Splatting rendering approach~\cite{DBLP:journals/tog/KerblKLD23} at any novel viewpoint. It's a differentiable tile-based rasterization that allows fast $\alpha$-blending of anisotropic splats and fast backward pass by tracking accumulated $\alpha$ values, ensuring that our pipeline can be trained end-to-end more efficiently with higher resolution images and less GPU memory cost.

\subsection{Reconstruction from Single-View Images}\label{subsec:rec}
In this section, we detail the process of learning the Triplane-Gaussian representation from a single image.  Drawing inspiration from recent breakthroughs in 3D reconstruction using transformers~\cite{DBLP:journals/corr/abs-2311-04400}, our approach employs two transformer-based decoders, each tailored for reconstructing a point cloud $P$ and a triplane $F$ from input images.
The adoption of a scalable transformer architecture substantially enhances the model's capability and generalization potential.
Features from the input image, along with camera features, are integrated into the transformer blocks of both the point and triplane decoders via cross-attention
Moreover, we harness local features projected from the input image to enhance both the point cloud up-sampling process and the 3D Gaussian decoder. This strategy significantly improves initial point cloud reconstruction and facilitates more accurate novel-view synthesis.

\vspace{-15pt}
\paragraph{Image Encoder.}
We first leverage a pre-trained ViT-based model~\cite{DBLP:conf/iclr/DosovitskiyB0WZ21} (e.g., DINOv2~\cite{dinov2}) to obtain patch-wise feature tokens from the input image.
In order to better utilize camera information in our pipeline, we implement an adaptive layer norm (adaLN) to modulate the DINOv2 features with the camera features, which is similar to \cite{DBLP:journals/corr/abs-2311-04400,DBLP:conf/cvpr/Wu0MFG23}.
Specifically, we project the camera parameters, which are the concatenation of flattened camera extrinsic matrix $T \in R^{4\times 4}$ and normalized intrinsic matrix $K \in R^{3\times 3}$, to high dimensional camera features $f_c \in R^{25}$ aligned with the DINOv2 feature dimension.
Finally, we implement an MLP to predict the scale and shift based on the camera features after each normalization layer.
These camera modulation layers are trained to make image features aware of the observation viewpoint, thereby better guiding both the point cloud network and the triplane network.

\vspace{-15pt}
\paragraph{Transformer Backbone.}
In our framework, we use a set of feature tokens $\{f_i\}_{p}$ and $\{f_i\}_{t}$ for the latent features of two different 3D representations, i.e., points and triplane, respectively.
The latent tokens are initialized from learnable positional embeddings, and fed to the transformer blocks.
Similar to other transformer architecture designs~\cite{jaegle2021perceiver}, each transformer block comprises a self-attention layer, a cross-attention layer, and a feed-forward layer.
The viewpoint-augmented image tokens guide the two decoders via cross-attention, respectively.

\vspace{-15pt}
\paragraph{Point Cloud Decoder.}
The point cloud decoder provides the coarse geometry of the object, 
where 3D Gaussians can be produced based on the coordinates of the points.
We employ a 6-layer transformer backbone to decode a point cloud using image conditions
from a fixed number of learnable positional embeddings,
i.e., point cloud tokens.
We treat each latent token as a point, which is similar to Point-E~\cite{DBLP:journals/corr/abs-2212-08751}.
However, our approach operates in a feed-forward manner, where the point cloud can be predicted directly from the final token,
as opposed to the diffusion process employed in Point-E~\cite{DBLP:journals/corr/abs-2212-08751}.
Due to limited computation and memory resources, 
we only decode a coarse point cloud with 2048 points in this step.

\vspace{-10pt}
\paragraph{Point Upsampling with Projection-Aware Conditioning.}
Since the quality of novel view synthesis from Gaussian Splatting is highly influenced by the number of Gaussians, 
it's not enough to generate 3D Gaussians from such a low-resolution point cloud.
Therefore, we adopt 
a lifting module with two-step Snowflake point deconvolution (SPD)~\cite{xiang2023SPD,xiang2021snowflakenet} to 
densify the point clouds from 2048 points to 16384 points.
Snowflake utilizes a global shape code extracted from 
the input point cloud along with point features, to predict 
the point displacements for the upsampled point cloud.
In order to reconstruct the detailed geometry guided by the input image, 
we integrate local image features into the shape code of SPD from the given camera pose
by Projection-Aware Conditioning~\cite{DBLP:conf/cvpr/Melas-Kyriazi0V23}.
Such projection enables the generation of high-resolution point clouds that are well-aligned with the input image.

\vspace{-10pt}
\paragraph{Triplane Decoder with Geometry-Aware Encoding.}
The triplane decoder outputs the implicit feature field based on the image and initial point cloud, from which 3D Gaussian properties will be decoded by position queries.
It shares a similar transformer architecture to decode the features from learnable positional embedding by image tokens but with deeper layers (we employ 10 layers for the triplane decoder).
We also encode the point cloud into the initial learnable positional embeddings, 
resulting in better geometry awareness.
More specifically, the points are also augmented with projection features from the input image as in point cloud upsampling. 
The feature-rich points are then fed into a shallow PointNet~\cite{DBLP:conf/cvpr/QiSMG17} with local pooling~\cite{DBLP:conf/eccv/PengNMP020}, and then we perform an orthographic projection onto the three axis-aligned planes.
The features projected to the same token are average-pooled and added with the learnable positional embedding.

\begin{figure*}
    \centering
    \includegraphics[width=\linewidth]{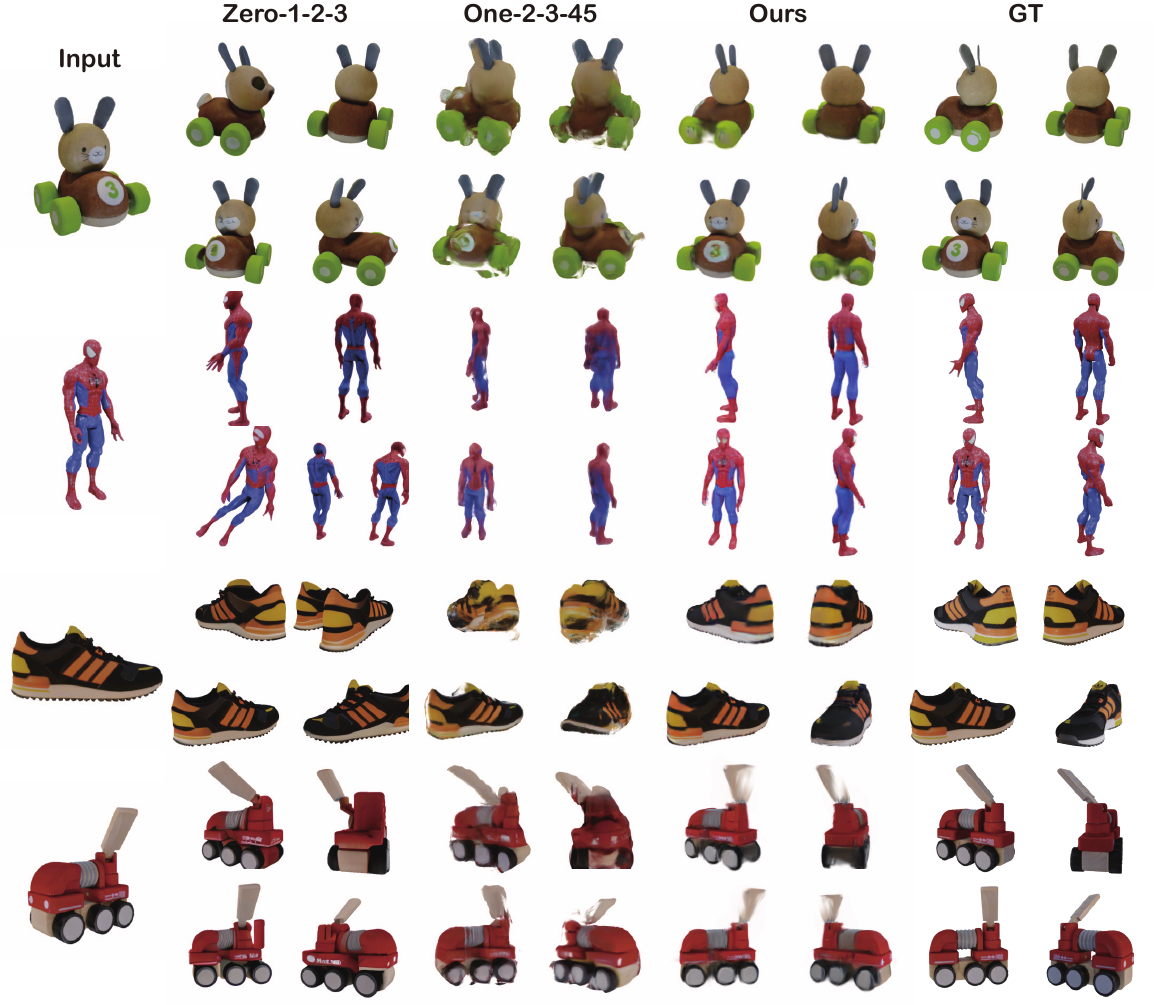}
    \caption{\textbf{Qualitative comparisons of novel view synthesis from reconstructed object between our method and other baselines on the GSO dataset.} Our approach achieves both quality and consistency across different novel views.}
    \label{fig:nvs}
    \vspace{-3mm}
\end{figure*}

\subsection{Training}\label{subsec:train}
We train the full pipeline by using 2D rendering loss along with 3D point cloud supervision:
\begin{align}
    \mathcal{L}&=\lambda_{c}\mathcal{L}_{\text{CD}} + \lambda_{e}\mathcal{L}_{\text{EMD}} + \\
    &\frac{1}{N}\sum_{i=1}^{N}(\mathcal{L}_{\text{MSE}} +
    \lambda_{m}\mathcal{L}_{\text{MASK}} + \lambda_{s}\mathcal{L}_{\text{SSIM}} + \lambda_{l}\mathcal{L}_{\text{LPIPS}})
\end{align}

where $\mathcal{L}_{\mathrm{CD}}$ is the Chamfer Distance (CD) and $\mathcal{L}_{\mathrm{EMD}}$ is the Earth Mover’s Distance (EMD).
For training the triplane decoder and 3D Gaussian decoder, we apply the rendering losses, including a pixel-wise MSE loss $\mathcal{L}_{\text{MSE}}=||I-\hat{I}||^2_2$, a mask loss $\mathcal{L}_{\text{MASK}}=||M-\hat{M}||^2_2$, a SSIM loss $\mathcal{L}_{\text{SSIM}}$~\cite{DBLP:journals/tip/WangBSS04} and a perceptual loss $\mathcal{L}_{\text{LPIPS}}$~\cite{DBLP:conf/cvpr/ZhangIESW18}.



\vspace{-3pt}
\section{Experiments}
In this section, we first give the implementation details, comparison baselines, and evaluation protocols.
Then we conduct two types of experiments to evaluate our method and other baseline methods for both quantitative and qualitative performance on geometry reconstruction and novel view synthesis.

\vspace{-5pt}
\subsection{Implementation Details}
\vspace{-5pt}
We utilize a pre-trained DINOv2~\cite{dinov2} as our image encoder, which generates 768-dimension feature tokens with a patch size of 14.
The point decoder is a 6-layer transformer with hidden dimension 512 and the triplane is a 10-layer transformer with hidden dimension 512.
The positional embeddings of the point decoder consist of 2048 tokens, each with 512 dimensions, corresponding to 2048 points.
The positional embeddings of the triplane decoder comprise three $32\times 32$ tokens, each with 512 dimensions, representing three axis-aligned planes.
To ensure better convergence during training, we divide the training process into two stages. 
Initially, we train the point decoder using $\lambda_{c}=10$ and $\lambda_{e}=10$ only with CD loss and EMD loss, enabling the network to deliver high-quality geometry.
Subsequently, we train the triplane decoder and the Gaussian decoder based on the input image and initial points, with the point decoder frozen.
We set the weights of the losses with $\lambda_{m}=1, \lambda_{s}=1, \lambda_{l}=2, \lambda_{c}=10, \lambda_{e}=0$ in this stage.

\begin{figure}
    \centering
    \includegraphics[width=\linewidth]{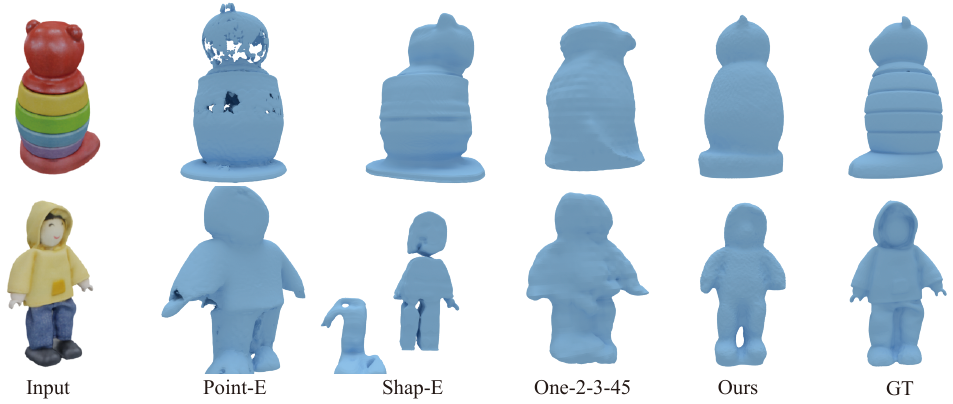}
    \caption{\textbf{Qualitative comparisons of geometry reconstruction from a single image between our method and other baselines on the GSO dataset.} The mesh of Shap-E and One-2-3-45 are extracted from implicit fields by Marching Cubes~\cite{DBLP:conf/siggraph/LorensenC87}, while the mesh of Point-E and ours are reconstructed from point cloud using a pre-trained regression-based model provided by \cite{DBLP:journals/corr/abs-2212-08751}.}
    \label{fig:mesh}
\end{figure}

\vspace{-5pt}
\subsection{Baselines, Dataset, and Metrics}
\paragraph{Baselines.}
We compare our method with the previous state-of-the-art single-image reconstruction and generation methods.
There are three kinds of them: (1) native 3D generative models, including Point-E~\cite{DBLP:journals/corr/abs-2212-08751} and Shap-E~\cite{DBLP:journals/corr/abs-2305-02463}; (2) 2D diffusion model, i.e., Zero-1-2-3~\cite{DBLP:journals/corr/abs-2303-11328} and (3) feed-forward model based on 2D diffusion model output, i.e., One-2-3-45~\cite{liu2023one}.
Point-E~\cite{DBLP:journals/corr/abs-2212-08751} and Shap-E~\cite{DBLP:journals/corr/abs-2305-02463} are trained on a large-scale internal 3D dataset to produce point cloud and implicit representation (NeRF) from single images. Zero-1-2-3~\cite{DBLP:journals/corr/abs-2303-11328} is a 2D generative model to utilize view-conditioned diffusion for novel view synthesis. 
One-2-3-45~\cite{liu2023one} trains a robust multi-view reconstruction model which takes multi-view images generated from a 2D diffusion model (e.g., Zero-1-2-3). It directly builds an SDF field by SparseNeuS~\cite{DBLP:conf/eccv/LongLWKW22}, achieving state-of-the-art performance. 

\vspace{-10pt}
\paragraph{Dataset.}
Following One-2-3-45~\cite{liu2023one}, we train our model on the Objaverse-LVIS dataset~\cite{DBLP:conf/cvpr/DeitkeSSWMVSEKF23}, which contains 46K 3D models in 1,156 categories.
We use Blender to render ground-truth RGB and depth with a circle path for an object.
For Evaluation, we adopt the Google Scanned Objects (GSO) dataset~\cite{DBLP:conf/icra/DownsFKKHRMV22}, which includes a wide variety of high-quality scanned household items, to evaluate the performance of our method and other baselines.
We randomly choose 100 shapes and render a single image per shape for evaluation. 

\vspace{-10pt}
\paragraph{Metrics.}
Following prior works, we adopt commonly used reconstruction metrics, including Chamfer Distance (CD) and Volume IoU, between ground truth and reconstruction, to evaluate the quality of geometry reconstruction from single-view images.
Furthermore, we also adopt PSNR, SSIM, and LPIPS metrics, which cover different aspects of image similarity, to evaluate the novel view synthesis performance.

\subsection{Single View Reconstruction}
We conduct the evaluation quality of geometry reconstruction from single-view images with other baselines, including Point-E~\cite{DBLP:journals/corr/abs-2212-08751}, Shap-E~\cite{DBLP:journals/corr/abs-2305-02463} and One-2-3-45~\cite{liu2023one}.
Table~\ref{tab:gso_geometry} demonstrates the quantitative results on CD and Volume IoU, and Figure~\ref{fig:mesh} illustrates some visual comparisons. 
Point-E tends to produce holes in the reconstructed meshes due to the sparsity of the generated point cloud.
Meanwhile, Shap-E is susceptible to collapsing during the generation process, resulting in an unpredictable outcome. 
The geometric output generated by One-2-3-4-5 \cite{liu2023one} is characterized by a coarse representation, lacking intricate details.
In contrast, our method reconstructs geometry based on the explicit point clouds, 
which can preserve finer details and maintain consistency with the ground truth.


\begin{table}[]
    \centering
    \begin{tabular}{c|cccc}
        \toprule[1pt]
        & CD $\downarrow$ &  IoU $\uparrow$ & Time (s) $\downarrow$ \\
        \toprule[1pt]  
        Point-E & 68.56 &  0.181 &  78\\ 
        Shape-E & 79.84 &  0.218 & 27\\ 
        One-2-3-45 & 75.56  & 0.338 & 40 \\ 
        Ours & \textbf{21.04} &  \textbf{0.401} & \textbf{0.14} \\ 
        \bottomrule[1pt]
    \end{tabular}
    \caption{\textbf{Quantitative Comparison for single view 3D reconstruction on the GSO dataset},
    in terms of Chamfer Distance $\times 10^{-3}$, Volume IoU and runtime efficiency.
    }
    \label{tab:gso_geometry}
\end{table}

\begin{table}[]
    \centering
    \begin{tabular}{c|ccccc}
        \toprule[1pt]
        & PSNR $\uparrow$ & SSIM $\uparrow$ & LPIPS $\downarrow$ & Time $\downarrow$ \\
        \toprule[1pt]
        One-2-3-45 & 15.55 & 0.76 & 0.25 & 14 \\
        Zero-1-2-3 &   17.57 & 0.78 & 0.19 & 1.70 \\
        Ours & \textbf{23.15} & \textbf{0.87} & \textbf{0.13} & \textbf{0.003} \\
        \bottomrule[1pt]
    \end{tabular}
    \caption{\textbf{Quantitative comparison on novel-view synthesis from single images on the GSO dataset}, in terms of PSNR, SSIM, LPIPS, and runtime efficiency.}
    \label{tab:gso_nvs}
\end{table}

\subsection{Novel View Synthesis}
In this subsection, we evaluate the quality of novel view synthesis in comparison with baselines, including One-2345 and Zero-1-2-3. The quantitative and qualitative results are shown in Table~\ref{tab:gso_nvs} and Figure~\ref{fig:nvs}, respectively.
While Zero-1-2-3 can generate plausible results from a novel viewpoint, it doesn't ensure consistency between different views, and can even yield completely unreasonable results. 
For example, it may generate two different instances of Spider-Man and shoes when the input only contains one of each.
Although One-2-3-45 is primarily designed for geometry reconstruction, we utilize it to render the novel view image from its generated feature volume.
Due to its inherent geometry constraint in 3D volume, the rendering images are consistent across different views.
However, its reliance on images generated by Zero-1-2-3 limits its ability to produce high-quality novel view synthesis.
In contrast, Our method delivers both high quality and consistency, thereby attaining superior performance.
Additionally, by leveraging the transformer architecture and local feature projection, our model exhibits robust generalization to unseen objects while preserving intricate textures.

\begin{table}[]
    \centering
    \resizebox{0.45\textwidth}{!}{
    \begin{tabular}{c|ccc}
        \toprule[1pt]
        & PSNR $\uparrow$ & SSIM $\uparrow$ & LPIPS $\downarrow$ \\ 
        \toprule[1pt]
        3DG & 18.56 & 0.80 & 0.20 \\ 
        Triplane-NeRF & 21.85 & 0.84 & 0.15 \\ 
        Triplane-Gaussian & \textbf{23.15} & \textbf{0.87} & \textbf{0.13} \\ 
        \bottomrule[1pt]
    \end{tabular}
    }
    \caption{\textbf{Quantitative comparison between different representations for novel view synthesis on GSO.}}
    \label{tab:gso_ab}
\end{table}

\begin{table}[]
    \centering
    \begin{tabular}{c|cccc}
        \toprule[1pt]
         & CD $\downarrow$ &  IoU $\uparrow$  \\
        \toprule[1pt]  
        PCL tokens       & 22.16  & 0.385 \\ 
        Global image feature  & 22.22  & \textbf{0.401} \\ 
        Projection-aware condition & \textbf{21.04} & \textbf{0.401}  \\ 
        \bottomrule[1pt]
    \end{tabular}
    \caption{\textbf{Quantitative Comparison of different conditions for point cloud up-sampling in geometry reconstruction}, in terms of Chamfer Distance $\times 10^{-3}$ and Volume IoU.
    }
    \label{tab:ab_upsample}
\end{table}

\subsection{Runtime Efficiency}
We also evaluate the runtime efficiency of our method in comparison with other baseline approaches.
Table~\ref{tab:gso_geometry} and Table~\ref{tab:gso_nvs} demonstrate the runtime of reconstruction and rendering of each baseline, respectively.
We evaluate the runtime of Zero-1-2-3's diffusion sampling process due to its lack of rendering process from 3D representation.
We evaluate the volume rendering of One-2-3-45 from its generated implicit feature volume rather than the generated mesh.
We can find that our method has achieved significant improvements in speed for both reconstruction and rendering processes compared to other baselines, benefiting from feed-forward fashion and efficient rasterization.

\subsection{Ablation Study}\label{subsec:ab}
\begin{figure}
    \centering
    \includegraphics[width=\linewidth]{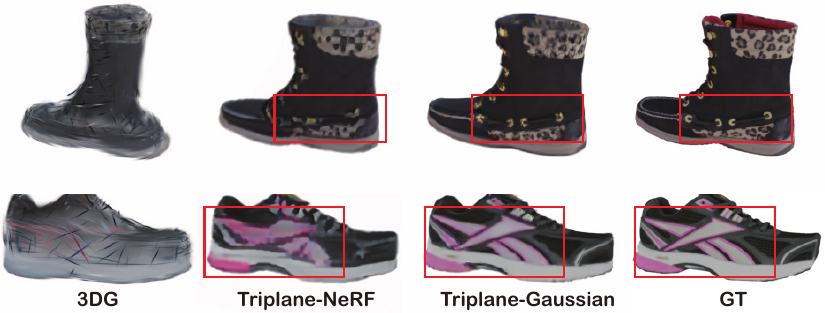}
    \caption{\textbf{Qualitative comparison with 3DG and Triplane-NeRF.} More accurate recovery of detailed textures by our hybrid representation is highlighted within the red box. }
    \label{fig:ab}
\end{figure}

\paragraph{3D representation.}
To evaluate the effectiveness of the design of our Triplane-Gaussian design, we conducted an experiment that compared it with two ablation settings, including (1) native generalizable 3D Gaussians (3DG) and (2) Triplane-NeRF.
Table~\ref{tab:gso_ab} and Figure~\ref{fig:ab} present the quantitative and qualitative comparisons among these methods.
The native generalizable 3D Gaussians, which are directly decoded from latent tokens like points, exhibit the poorest performance according to all metrics. 
Our Triplane-Gaussian, leveraging the projection-aware condition with explicit geometry, excels in producing more detailed texture compared to Triplane-NeRF, as illustrated in the red box of Figure~\ref{fig:ab}, and achieves the best quantitative results.
Furthermore, once the 3D Gaussians are decoded, our rendering process proves to be faster than that of Triplane-NeRF.
This efficiency arises because our method doesn't require decoding features from the triplane for each sampling point along the ray marching path.
As the rendering resolution increases, the superiority of Triplane-Gaussian becomes increasingly evident.


\begin{table}[t]
    \centering
    \begin{tabular}{c|cc|ccc}
    \toprule[1pt]
        & P.C. & G.E. & PSNR $\uparrow$ & SSIM $\uparrow$ & LPIPS $\downarrow$ \\
    \midrule[1pt]
       a & \ding{55} & \ding{55} & 22.04 & 0.85 & 0.16 \\
       b & \ding{51} & \ding{55} & 22.47 & 0.86 & 0.15 \\
       c & \ding{55} & \ding{51} & 22.76 & 0.86 & 0.14 \\
       d & \ding{51} & \ding{51} & \textbf{23.15} & \textbf{0.87} & \textbf{0.13} \\
    \bottomrule[1pt]
    \end{tabular}
    \caption{\textbf{Quantitative effect of projection-aware condition and geometry-aware encoding to novel view synthesis.}}
    \label{tab:ab_method}
\end{table}

\begin{figure*}
    \centering
    \includegraphics[width=\linewidth]{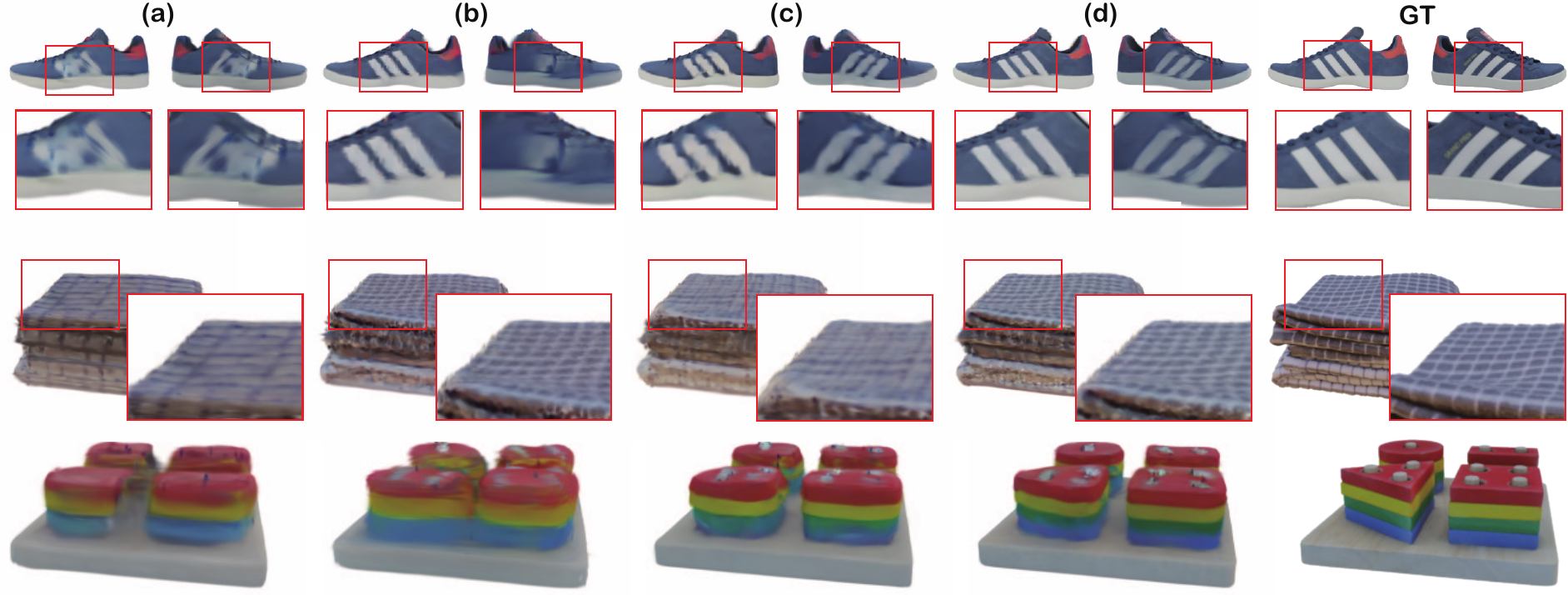}
    \caption{\textbf{Qualitative effect of projection-aware condition and geometry-aware encoding.} The (a-d) are corresponding with (a-d) in Table~\ref{tab:ab_method}.}
    \label{fig:ab_method}
\end{figure*}

\vspace{-20pt}|
\paragraph{Projection-aware condition.}
\zzx{
We evaluate the effect of projection-aware condition on both point cloud reconstruction and novel view synthesis.
Firstly,} to investigate the impact of different shape codes 
within the point upsampling module, we conduct experiments with two ablation shape code settings, including (1) the \textbf{P}oint \textbf{CL}oud (PCL) tokens from the point cloud decoder and (2) global image features from DINOv2.
As shown in Table \ref{tab:ab_upsample}, image-based shape codes achieve higher IoU compared to merely PCL tokens,
which demonstrates the importance of incorporating structural information from the visual modality. Moreover, the projection-aware condition outperforms the other two code methods in terms of CD, suggesting that the local patterns of the image are effectively transferred to the structure of the point cloud.

\zzx{
We also assess the impact of the projection-aware condition on the 3D Gaussian Decoder for novel view synthesis. Table~\ref{tab:ab_method} demonstrates that using projection-aware condition (P.C.) improves the quality of novel view synthesis across all metrics.
In figure~\ref{fig:ab_method}, two examples illustrate its visual effects.
The use of P.C. (Figure~\ref{fig:ab_method} (b) and (d)) significantly enhances the rendering image with sharper and more detailed texture on the same side as the input view. For example, in the first row, the left image of the shoe, and in the second row, the complex checkered texture of the coverlet.

\paragraph{Geometry-aware encoding.}
While the projection-aware condition improves rendering quality on the same side as the input view, achieving good texture on the backside remains challenging (see the first row of Figure~\ref{fig:ab_method} (b)).
Introducing geometry-aware encoding (G.E.) enhances texture results, as evident in Figure~\ref{fig:ab_method} (c) and (d), particularly on the backside (right image of the shoe). 
Geometry-aware encoding enables the triplane decoder to predict backside texture with a shape prior, considering that shoes often have a similar texture on both sides.
Since our point cloud decoder and triplane decoder are trained separately, the implicit field is sometimes not well-aligned with the explicit point cloud for complex geometry in the absence of geometry-aware encoding. This leads to blurry and non-sharp rendering results, as observed in the last row of Figure~\ref{fig:ab_method} (a) and (b). The geometry-aware encoding enhances the alignment between them, resulting in high-quality novel view synthesis.
Table~\ref{tab:ab_method} also demonstrates the quantitative improvement achieved by geometry-aware encoding in novel view synthesis results.
}


\subsection{Limitations}
The rendering quality of the 3D Gaussians largely depends on the initial geometry. If the predicted point cloud deviates severely from the ground truth, it becomes challenging for the 3D Gaussians to recover the missing regions.
Additionally, since our model is not a probabilistic model, the backside tends to be blurry.
Our model also relies on the camera parameters, which will restrict its applications.
\vspace{-5pt}
\section{Conclusion}
In this paper, we present \shortname, a new framework for 3D reconstruction from a single image with only a feed-forward inference.
When presented with an image of an object, our approach constructs a hybrid Triplane-Gaussian representation using two transformer decoders - a point cloud decoder and a triplane decoder.
This is followed by a 3D Gaussian decoder to generate Gaussian properties, e.g. opacity and spherical harmonics.
To improve the consistency with the input observation, we further augment the generation process of these two representations by the application of projection-aware conditioning and geometry-aware encoding.
Experiment results demonstrate that our method not only achieves high-quality geometry reconstruction and novel view synthesis, but also maintains a fast reconstruction and rendering speed.

{
    \small
    \bibliographystyle{ieeenat_fullname}
    \bibliography{main}
}


\end{document}


\maketitlesupplementary





\section{Qualitative Comparison on Real-World Images}
\begin{figure*}[t]
    \centering
    \includegraphics[width=\linewidth]{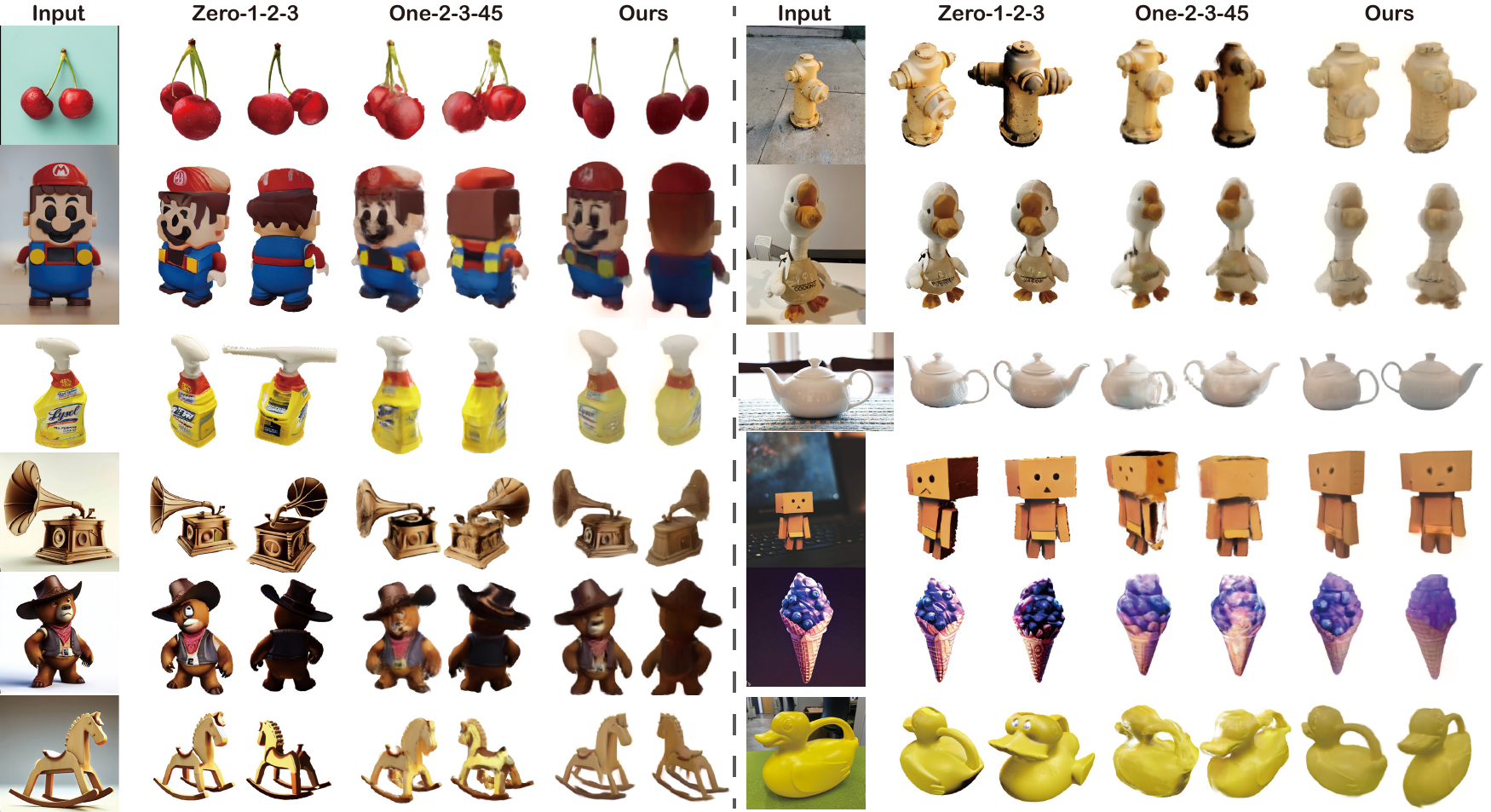}
    \caption{Qualitative comparison with Zero-1-2-3 and One-2-3-45 on real-world images.}
    \label{fig:supp_real}
\end{figure*}

We also provide a qualitative evaluation of novel view synthesis on real-world images, comparing our method with One-2-3-45~\cite{liu2023one} and Zero-1-2-3~\cite{DBLP:journals/corr/abs-2303-11328}. 
The real-world images used in our evaluation are sourced from One-2-3-45\footnote{https://github.com/One-2-3-45/One-2-3-45/tree/master/demo/demo\_examples}, captured in real-world or generated from 2D image generative model (e.g. DALL-E~\cite{DBLP:conf/icml/RameshPGGVRCS21}).
We leverage the elevation estimation module from One-2-3-45 to estimate the elevation of the input and assume the azimuth of the input is $0$.
In Figure~\ref{fig:supp_real}, a qualitative comparison between ours and baselines is presented.
Similar to the observed experimental phenomena on the GSO~\cite{DBLP:conf/icra/DownsFKKHRMV22}, 
Zero-1-2-3 faces challenges in maintaining consistency across different novel views. 
While One-2-3-45 demonstrates proficiency in geometry recovery, it exhibits limitations in the quality of rendering images.
In contrast, our method not only achieves high-quality novel view synthesis but also maintains consistency across views, closely adhering to the input image.
Despite being trained on a synthetic dataset, our method generalizes well to real-world images. 
Moreover, it exhibits robustness to estimated camera poses.

\section{More Ablation Analysis}
\paragraph{Projection-aware condition and Geometry-aware encoding.}
In addition to the evaluation of projection-aware condition (P.C.) to point cloud generation in our paper, we conduct an ablation experiment here to demonstrate its influence on novel view synthesis.
Furthermore, we assess the effect of geometry-aware encoding (G.E.) before the triplane decoder.
Table~\ref{tab:supp_ab} presents the quantitative results under different ablative experiment settings, including (a) training without projection-aware condition and geometry-aware encoding; (b) training only with projection-aware condition, and (c) training with both components.
It shows that utilizing both components leads to an improvement in novel view synthesis quality across all metrics.

\begin{figure}[t]
    \centering
    \includegraphics[width=\linewidth]{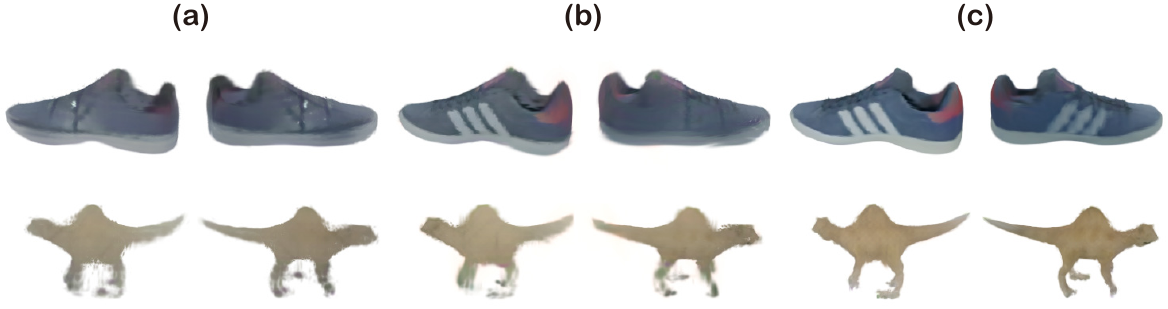}
    \caption{Qualitative effect of projection-aware condition and geometry-aware encoding to novel view synthesis.}
    \label{fig:supp_ab}
\end{figure}

Figure~\ref{fig:supp_ab} illustrates two examples to show the effects of the projection-aware condition (P.C.) and geometry-aware encoding (G.E.).
In comparison to the base model (a) without P.C. and G.E., the use of P.C. (b) significantly enhances the texture on the same side of the input view (left image of shoe in (b)).
However, details on the opposite side are still lacking (right image of shoe in (b)).
Upon applying G.E., our full model (c) exhibits improved texture on the opposite side (right image of shoe in (c)), primarily attributed to the geometry-aware encoding.
Geometry-aware encoding enables the triplane decoder to predict the opposite-side texture with a shape prior. 
For example, the shoe often has a similar texture on the two sides.
Since our point cloud decoder and triplane decoder are trained separately, the implicit triplane is sometimes not well-aligned to the explicit point cloud for complex geometry, leading to blurry and non-sharp rendering results (the second row of Figure~\ref{fig:supp_ab} (a) and (b)).
The geometry-aware encoding enhances the alignment between the point cloud and triplane, resulting in high-quality novel view synthesis.

\begin{table}[t]
    \centering
    \begin{tabular}{c|cc|ccc}
    \toprule[1pt]
        & P.C. & G.E. & PSNR $\uparrow$ & SSIM $\uparrow$ & LPIPS $\downarrow$ \\
    \midrule[1pt]
       a & \ding{55} & \ding{55} & 21.00 & 0.84 & 0.18 \\
       b & \ding{51} & \ding{55} & 21.39 & 0.85 & 0.17\\
       c & \ding{51} & \ding{51} & \textbf{21.82} & \textbf{0.86} & \textbf{0.15} \\
    \bottomrule[1pt]
    \end{tabular}
    \caption{Quantitative effect of projection-aware condition and geometry-aware encoding to novel view synthesis.}
    \label{tab:supp_ab}
\end{table}

\begin{table}[t]
    \centering
    \begin{tabular}{c|ccc}
    \toprule[1pt]
        Num of Points & PSNR $\uparrow$ & SSIM $\uparrow$ & LPIPS $\downarrow$ \\
    \midrule[1pt]
        2048 & 20.62 & 0.84 & 0.20 \\
        16384 & \textbf{21.82} & \textbf{0.86} & \textbf{0.15} \\
    \bottomrule[1pt]
    \end{tabular}
    \caption{Quantitative effect of point upsampling to novel view synthesis.}
    \label{tab:supp_up}
\end{table}

\begin{figure}[h]
    \centering
    \includegraphics[width=\linewidth]{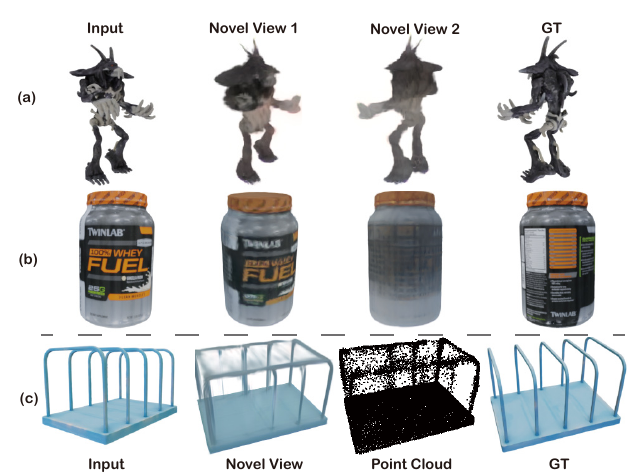}
    \caption{Failure cases.}
    \label{fig:supp_failure}
    \vspace{-10pt}
\end{figure}


\paragraph{Point Upsampling.}
The number of 3D Gaussians plays a crucial role in determining the quality of 3D reconstruction.
Even in a per-shape optimization scenario, as did in \cite{DBLP:journals/tog/KerblKLD23}, a reduction in the number of 3D Gaussians can result in poorer performance.
Given our adoption of a transformer architecture for point cloud generation, this decoder is constrained by computation and memory resources, making it unable to directly output high-resolution point cloud.
To achieve a dense point cloud, we leverage a point upsampling module adapted from Snowflake point deconvolution (SPD)~\cite{xiang2023SPD,xiang2021snowflakenet} to densify the point cloud from 2048 points to 16384 points.
Table~\ref{tab:supp_up} demonstrates a comparison of novel view synthesis quality between using point upsampling (with 16384 points) and not using point upsampling (with 2048 points).
The results indicate that the performance of novel view synthesis decreases with the reduction in the number of 3D Gaussians.

\section{More Qualitative results}
\begin{figure*}[h]
    \centering
    \includegraphics[width=\linewidth]{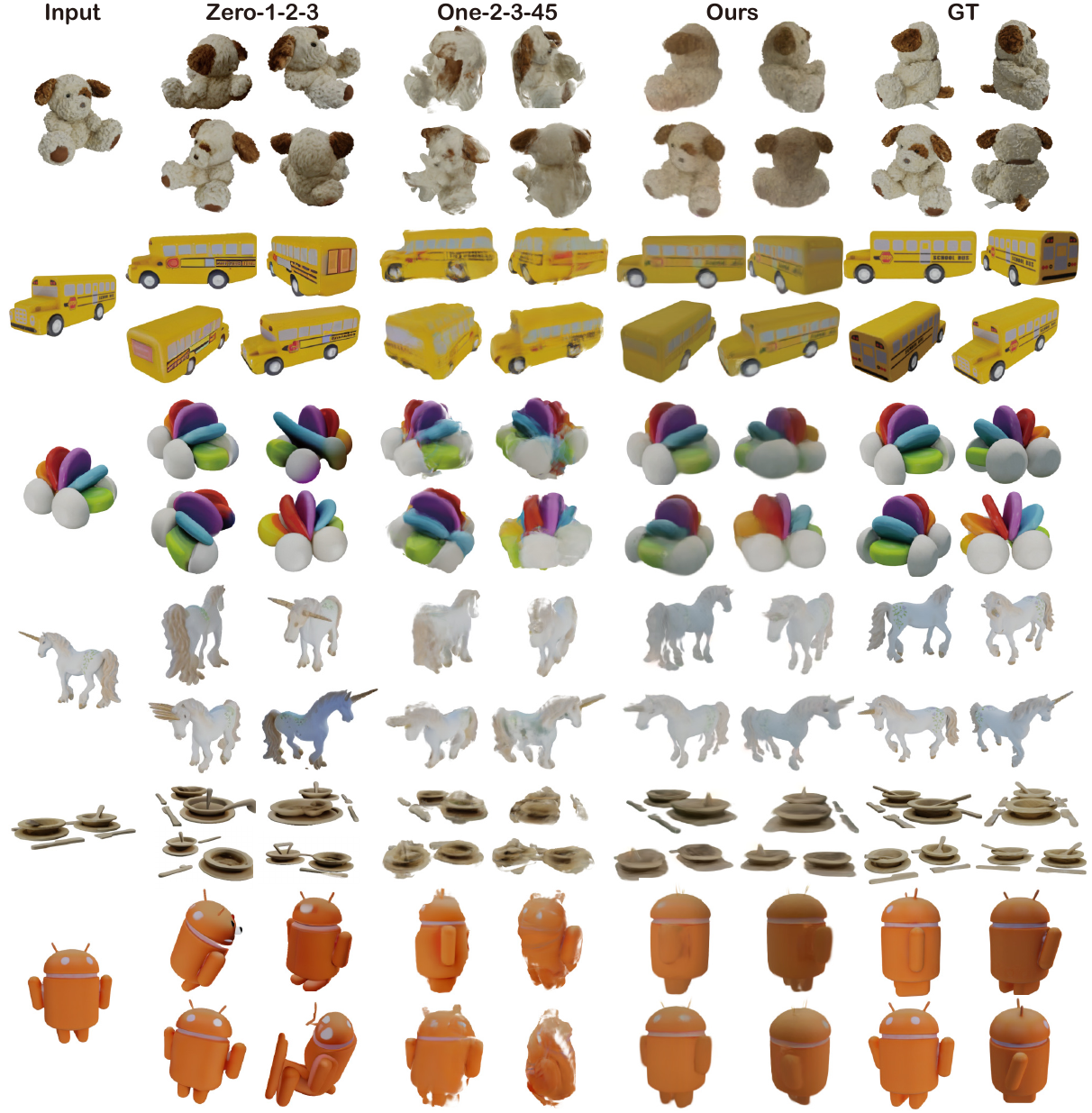}
    \caption{More qualitative comparison with Zero-1-2-3 and One-2-3-45 on GSO dataset.}
    \label{fig:supp_nvs}
\end{figure*}

We present additional visual comparisons between the two baselines and ours in Figure~\ref{fig:supp_nvs}.
Consistent with previous findings, Zero-1-2-3 struggles to maintain consistency while One-2-3-4-5 has relatively poor performance on novel view synthesis.

Furthermore, we have included an offline web demo in the supplementary materials to showcase the 360-degree visualizations of both the other baselines and our method.
Please refer to it for more details.

\section{Failure Cases and Limitations}
While our method has demonstrated effectiveness, there are still some limitations, as illustrated in Figure~\ref{fig:supp_failure}. 
As discussed in the paper, our regression method often struggles to ``imagine'' the backside (except it sometimes can guess the backside through shape symmetric prior as Figure~\ref{fig:supp_ab} (c) shown), resulting in blurry texture (see Figure~\ref{fig:supp_failure} (b)).
While our method can roughly reconstruct the geometry, it encounters challenges in accurately reconstructing complex action figures, as depicted in Figure~\ref{fig:supp_failure} (a).
Furthermore, Figure~\ref{fig:supp_failure} (c) illustrates that inaccurate point cloud estimation can adversely affect the accuracy of our 3D Gaussians.
To improve our method, the potential solutions could include: (1) designing a mechanism to facilitate feature interactions between the point cloud decoder and the triplane decoder and (2) exploring a diffusion model based on 3D Gaussian to achieve improved texture results, especially on the opposite side.

{
    \small
    \bibliographystyle{ieeenat_fullname}
    \bibliography{main}
}
